\documentclass{article} 
\usepackage{iclr2027_conference,times}

\iclrfinalcopy

\usepackage{hyperref}
\usepackage{url}

\usepackage{amsmath,amsfonts,bm}

\def\eqref#1{equation~\ref{#1}}

\def\1{\bm{1}}

\DeclareMathAlphabet{\mathsfit}{\encodingdefault}{\sfdefault}{m}{sl}
\SetMathAlphabet{\mathsfit}{bold}{\encodingdefault}{\sfdefault}{bx}{n}

\usepackage{amssymb}
\usepackage{amsthm}

\usepackage{booktabs}
\usepackage{graphicx}
\usepackage{array}
\usepackage{tabularx}
\usepackage{wrapfig}
\usepackage{float}
\usepackage{tcolorbox}
\usepackage[table]{xcolor}
\tcbuselibrary{skins,breakable}
\definecolor{purple}{RGB}{111,36,174}
\newtcolorbox{normalbox}[1]{
    enhanced,
    breakable,
    title={#1},
    colback=black!1.5,
    colframe=black!18,
    coltitle=black,
    fonttitle=\bfseries,
    colbacktitle=black!1.5,
    boxrule=0.45pt,
    titlerule=0pt,
    arc=0.8mm,
    left=3mm,
    right=3mm,
    top=0.6mm,
    bottom=1.8mm,
    toptitle=1.2mm,
    bottomtitle=0mm,
    before skip=5pt,
    after skip=5pt,
}
\newtcolorbox{keybox}[1]{
    enhanced,
    breakable,
    colback=purple!4,
    colframe=purple!4,
    boxrule=0pt,
    borderline west={3pt}{0pt}{purple},
    sharp corners,
    left=4mm,
    right=3mm,
    top=2mm,
    bottom=2mm,
    fontupper=\normalfont,
    before upper={\textbf{#1}\par\smallskip},
}
\usepackage{placeins}
\usepackage[font=small,skip=4pt]{caption}

\AtBeginDocument{%
  \setcounter{topnumber}{2}%
  \setcounter{bottomnumber}{1}%
  \setcounter{totalnumber}{3}%
}

\newcolumntype{R}{>{\raggedleft\arraybackslash}X}

\title{Verbalizing Multi-Token Concepts in LLMs}

\author{
Xijie Gong$^{*,\ddagger}$
\qquad
Zimeng Huang$^{*}$
\qquad
Tonghan Wang$^{\dagger}$\\
College of AI, Tsinghua University, Beijing, China\\
{\small $^{\dagger}$Correspondence to: \texttt{thw@mail.tsinghua.edu.cn}}
}

\begin{document}

\maketitle

\begingroup
\renewcommand{\thefootnote}{}
\footnotetext{$^{*}$Equal contribution. \quad $^{\ddagger}$Work done during an internship at Tsinghua University.}
\addtocounter{footnote}{-1}
\endgroup

\begin{abstract}
Lens methods inspect model computation by mapping intermediate activations to vocabulary tokens. Yet the concepts humans need to read out often span multiple tokens---entities, phrases, intermediate objects---making token-level readouts incomplete. Reliable multi-token readout with little model-specific preparation remains challenging. We introduce Concept Lens: token-level lens clues guide candidate concept search, then the model derives a representation for each candidate and scores it against the original activation. Across 2,400 multi-hop clozes on five LLMs (8B--70B), Concept Lens instantiated with J-lens and R-lens achieves average Rank@10 scores of 36.6\% and 54.5\%, respectively, compared with 21.7\% for Template Lens. 
Concept-swap interventions on derived concept representations shift model answers toward those associated with the replacement concepts. 
Further experiments show that Concept Lens can also reveal what a model recognizes along the way, beyond what appears in its final answer.
Our code is available at \url{https://github.com/XijieGo/c-lens}.
\end{abstract}

\vspace{-8pt}
\begin{figure}[!ht]
\centering
\includegraphics[width=0.98\linewidth]{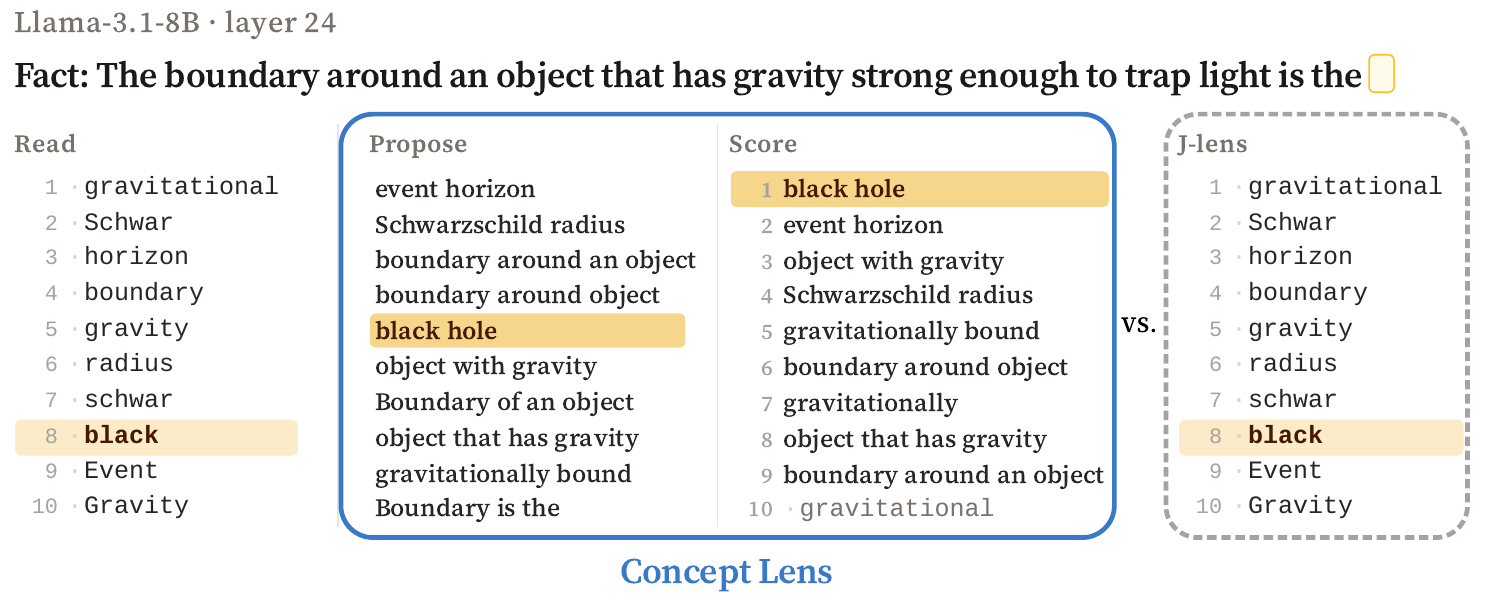}
\caption{\textbf{Our method extends J-lens readout from individual vocabulary tokens to multi-token concepts.} In this example, the model should infer \textit{black hole} before answering \textit{event horizon}. J-lens identifies the first-token clue \textit{black}, at rank 8. Concept Lens uses ranked first-token clues together with the source prompt to generate candidate multi-token concepts, then scores each candidate by deriving its representation and comparing against the original activation. This creates a unified readout space where candidate concepts and vocabulary tokens are ranked together. Our method assigns the highest score to the target concept, \textit{black hole}.}
\label{fig:teaser}
\end{figure}

\section{Introduction}
\label{sec:intro}

Large language models (LLMs) compute their outputs by transforming high-dimensional activations across layers. These activations carry information about intermediate stages of the computation. Lens methods make these activations more interpretable by mapping them into textual readout spaces~\citep{nostalgebraist2020logitlens,belrose2023tunedlens,gurnee2026verbalizable,nanda_review_2026,blank2026rlens}. These readouts provide a human-readable interface to inspect a model's internal computation and can reveal latent concepts that contribute to an answer yet remain unstated in the final output.

\begin{wrapfigure}{r}{0.40\textwidth}
\centering
\includegraphics[width=\linewidth]{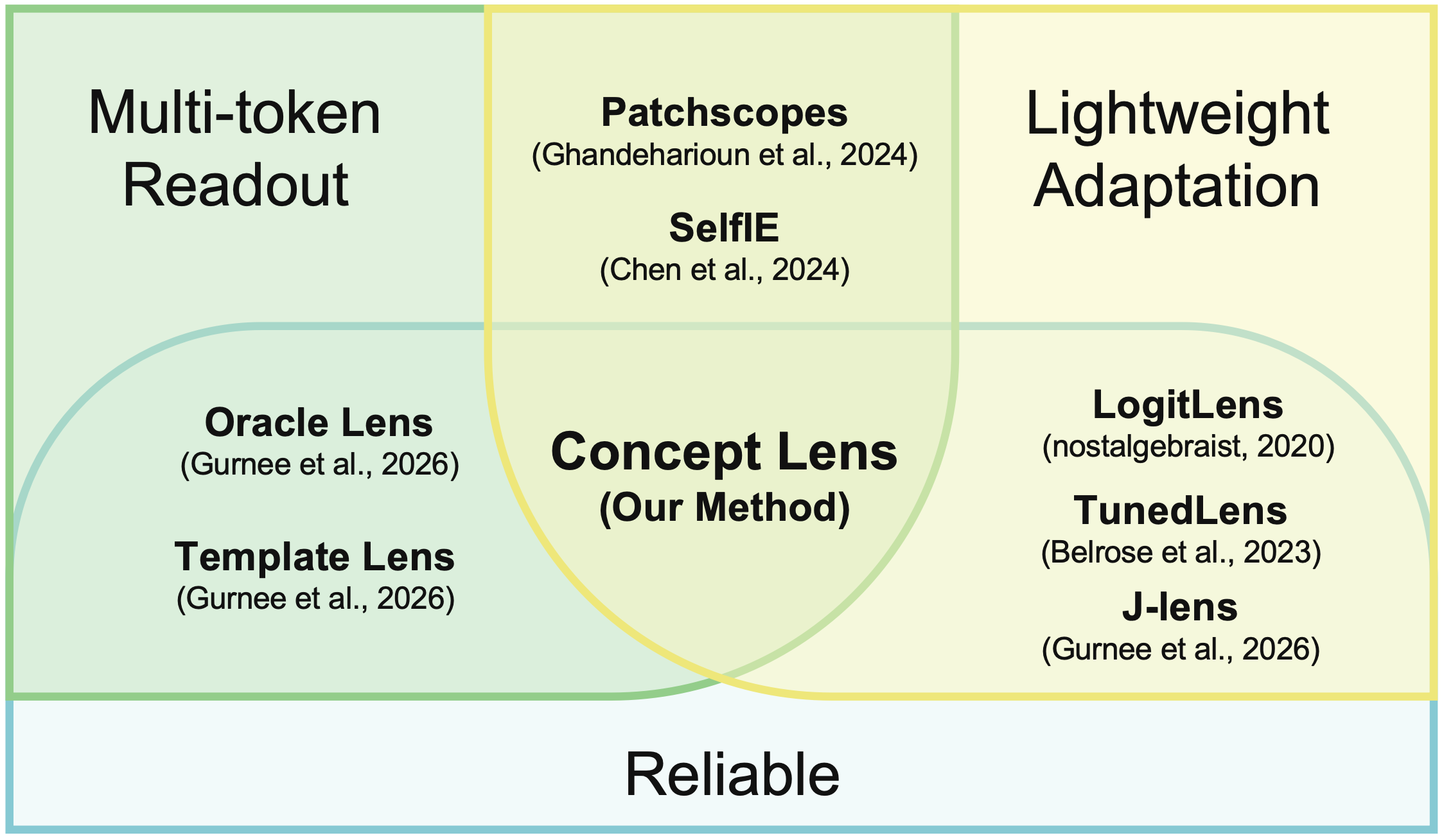}
\caption{Compared to existing lens methods, Concept Lens (our method) enables reliable multi-token readout with lightweight adaptation.}
\label{fig:concept-lens-comparison}
\vspace{-10pt}
\end{wrapfigure}
Existing lens methods, however, face a trade-off between expressiveness and preparation cost. Token-level methods are lightweight, but expose only individual tokens. Logit Lens~\citep{nostalgebraist2020logitlens}, Tuned Lens~\citep{belrose2023tunedlens}, J-lens~\citep{gurnee2026verbalizable,nanda_review_2026}, and R-lens~\citep{blank2026rlens}, for example, can surface the fragment \textit{black} even when the activation being read carries information about the complete concept \textit{black hole} (Figure~\ref{fig:teaser}). Sequence-level methods support complete concepts, but typically require more preparation: Template Lens relies on a predefined concept set, while Oracle Lens trains auxiliary models to decode free-form phrases~\citep{gurnee2026verbalizable}. Patchscopes and SelfIE generate multi-token descriptions with little preparation by asking a language model to describe an injected activation~\citep{pmlr-v235-ghandeharioun24a,pmlr-v235-chen24ao}. However, a plausible description may not accurately reflect the information in that activation. Faithful-Patchscopes shows that familiar word associations can override information carried by the injected activation during generation~\citep{gong2026faithfulpatchscopes}. This motivates our central question (Figure~\ref{fig:concept-lens-comparison}): \textbf{Can we obtain reliable readouts of multi-token concepts with little additional preparation for each model?}

We answer this question with a two-stage method \textbf{Concept Lens}. First,  \textbf{PROPOSE} uses token clues from an existing lens and the input context to generate candidate concepts. Second, \textbf{SCORE} places each candidate in a fixed carrier prompt and derives its representation from subsequent activations. We map these concept representations to the layer being read and rank concepts alongside vocabulary tokens by their agreement with the original activation. Each candidate's score is independent of its generation probability. This design allows us to directly test the reliability of the readout: concept representations can be intervened on to assess whether they have the expected causal effect on subsequent computation. Concept Lens requires only lightweight calibration on a new model.

Across five LLMs (8B--70B), Concept Lens reads out multi-token concepts and derives representations that support causal interventions. On 2,400 multi-hop cloze prompts, Concept Lens instantiated by J-lens and R-lens rank the target concept among the top 10 readout entries for 36.6\% and 54.5\% of prompts, respectively, versus 21.7\% for Template Lens. Further analyses show that, given the correct first token and the source prompt, the model infers the full target concept in 89.9\% of cases. In a controlled test where held-out single-token words are split into two tokens, Concept Lens derives the original word as its top choice in 98.0--99.8\% of cases. Finally, intervening on concept representations changes model answers in the predicted directions, providing causal evidence that representations exposed by the readout participate in the model's subsequent computation.

Moreover, Concept Lens reveals meaningful aspects of a model's internal computation that are not surfaced in its output. 
Consider two otherwise similar problems, one containing a contradiction and the other logically consistent (Figure~\ref{fig:applications}). The model produces the same answer---``Noor is the youngest''. Concept Lens, however, exposes an internal difference: when processing the contradictory input, the model represents \textit{contradiction between the statements}, whereas for the consistent input, it represents the task-relevant concept \textit{youngest person}. Thus, although the final answer shows no sign of conflict, our method reveals that it has already detected the contradiction. In other examples, Concept Lens highlights \textit{combinations formula} and \textit{permutations without repetition} in two counting problems with the same answer. In paired conversational contexts, \textit{humour or irony} ranks much higher when the same phrase is used sarcastically than sincerely. These results illustrate how Concept Lens reveals not only \textit{what} a model answers, but also \textit{what it recognizes along the way}.

\section{Preliminaries}
\label{sec:prelim}

\subsection{The J-lens Readout}
\label{sec:method-setup}

J-lens reads an intermediate hidden state through its effect on later model representations~\citep{gurnee2026verbalizable}. Let $h_{\ell,t}(x)$ denote the activation at layer $\ell$ and position $t$, $h_{\mathrm{final},t}$ the final-layer hidden state, and $W_U$ the model's unembedding matrix. Let $u_w\in\mathbb{R}^{d_{\mathrm{model}}}$ be the column form of row $w$ of $W_U$. We write the model's final RMS normalization as $N(z)=\Gamma z/r(z)$, where $\Gamma$ is the learned diagonal gain and $r(z)>0$ is the RMS scale. J-lens estimates the average Jacobian
\begin{equation}
J_\ell =
\mathbb{E}_{t,\,t'\ge t,\,x}
\left[
\frac{\partial h_{\mathrm{final},t'}(x)}
{\partial h_{\ell,t}(x)}
\right],
\label{eq:j-lens-jacobian}
\end{equation}
which captures how a perturbation to a layer-$\ell$ hidden state propagates, on average, to later final-layer representations. The resulting readout is
\begin{equation}
\mathrm{lens}(h_\ell)
=
\mathrm{softmax}
\left(
W_U\,N(J_\ell h_\ell)
\right).
\label{eq:j-lens-readout}
\end{equation}

Sorting $\mathrm{lens}(h_\ell)$ over the vocabulary gives the \emph{J-lens ranked list}. The J-lens authors interpret its top-ranked tokens as those the model is poised to verbalize. The pre-softmax logit for token $w$ is $\langle J_\ell^\top\Gamma^\top u_w,h_\ell\rangle/r(J_\ell h_\ell)$. Softmax preserves order, and the positive denominator is shared across tokens. The rank of $w$ is therefore determined by
\begin{equation}
\langle d_\ell(w), h_\ell\rangle,
\qquad
d_\ell(w) \triangleq J_\ell^\top\Gamma^\top u_w
\in \mathbb{R}^{d_{\mathrm{model}}},
\label{eq:token-representation}
\end{equation}
where we call $d_\ell(w)$ the \emph{token representation}. Each $d_\ell(w)$ specifies the direction in hidden-state space that favors token $w$ under the J-lens readout: a larger inner product gives $w$ a higher score.

\subsection{Multi-Token Concept Readout}
\label{sec:method-formulation}

Let $C=(w_1,\ldots,w_n)$ denote a multi-token concept with $n\ge2$, such as ``black hole.'' The unembedding matrix $W_U$ provides one row per vocabulary token, so J-lens scores the constituent tokens of $C$ individually. A complete concept readout must distinguish compositions such as ``black hole'' and ``black coffee,'' which share the token ``black.'' Thus, extending a token-level lens to multi-token readout requires not only the complete concept $C$, but also a representation specific to that concept, $v_C^{(\ell)}\in\mathbb{R}^{d_{\mathrm{model}}}$, at layer $\ell$. However, deriving a representation that faithfully captures the complete concept is nontrivial, since it must preserve the meaning of the full token sequence. A sum or mean of constituent token representations assigns the same vector to ``black hole'' and ``hole black.''

\subsection{Existing Extensions for Multi-Token Concepts}
\label{sec:prelim-extensions}

Existing approaches differ in how they provide $C$ and $v_C^{(\ell)}$. Patchscopes and SelfIE use language models to interpret their own activation as multi-token concepts with little preparation~\citep{pmlr-v235-ghandeharioun24a,pmlr-v235-chen24ao}. Causal verification of a decoded concept additionally requires deriving a corresponding representation. Template Lens and Oracle Lens provide such representations for readout and intervention, at substantial preparation cost~\citep{gurnee2026verbalizable}. Template Lens constructs representations for a predefined concept list, while Oracle Lens trains auxiliary models for concept generation and reconstruction. These approaches highlight a trade-off between lightweight adaptation and support for causal verification (Figure~\ref{fig:concept-lens-comparison}).

\begin{keybox}{Question}
Can we read multi-token concepts with lightweight adaptation while deriving representations for causal verification?
\end{keybox}

\section{Concept Lens}
\label{sec:method}

To address the question above, Concept Lens uses the model's own representations to connect complete concepts with the hidden state being inspected. We describe the method using J-lens.

\subsection{Proposing Candidate Concepts}
\label{sec:method-propose}

At source layer $\ell_s$, \textbf{PROPOSE} uses the J-lens ranking of $h_{\ell_s}$ to guide candidate search. We select tokens from this ranking as possible first tokens of complete concepts. To search beyond the highest-ranked entries, we partition a fixed prefix of the ranking into disjoint ranges and allocate a separate search budget to each range.
Each selected seed is placed in the fixed \emph{Scaffold Prompt}.

\begin{normalbox}{Scaffold Prompt}
\small\ttfamily
when model is reading the context below: \textbackslash n <cloze>.\\
it will think a concept: <seed>
\end{normalbox}

Here \texttt{<cloze>} is the source prompt and \texttt{<seed>} fixes the first token of the candidate. The scaffold can be tailored to the application to guide candidate generation. Within each rank range, we retain high-probability completions until termination or a fixed length limit, then merge and filter them to obtain a fixed candidate set.
Rank ranges, decoding budgets, and filtering rules are given in Appendix~\ref{app:implementation-propose}. \textbf{PROPOSE} probabilities guide candidate generation, and \textbf{SCORE} measures each candidate's agreement with the source hidden state independently of its proposal probability.

\subsection{Deriving and Scoring Candidate Representations}
\label{sec:method-score}

To score a candidate as a concept, \textbf{SCORE} constructs a representation from the model's processing of its full token sequence. We place $C$ in a fixed \emph{Carrier Prompt} and extract activation from a neutral factual continuation. Reading these later positions follows prior observations that concepts can remain readable in subsequent activation~\citep{gurnee2026verbalizable}. The candidate is the only variable part of the input, allowing its representation to be compared with different source states.

\begin{normalbox}{Carrier Prompt}
\small\ttfamily
Remember the following concept in your mind.\\
Concept:<$C$>. Fact: The capital of France is Paris.
\end{normalbox}

Let $\tau(C)$ denote the carrier input. We extract activation at selected positions $\mathcal{T}$ in the fixed continuation \texttt{The capital of France is Paris.}. After subtracting the position-specific background, we pool them as
\begin{equation}
v_C^{(\ell_c)}
=
\sum_{t\in\mathcal{T}}
\alpha_t
\left(h_{\ell_c,t}(\tau(C))-\mu_t\right),
\qquad
\alpha_t\geq0,\quad
\sum_{t\in\mathcal{T}}\alpha_t=1.
\label{eq:carrier}
\end{equation}
The weights $\alpha_t$ are calibrated independently on single-token concepts using agreement with J-lens token representations (Appendix~\ref{app:implementation-carrier}). Under the local shared-component model in Appendix~\ref{app:theory-recovery}, the pooled representation can be written as $v_C^{(\ell_c)}=\kappa_C g_C+\eta_C$, where $g_C$ is the concept-related direction and $\eta_C$ is the remaining pooled residual. If $\kappa_C>0$ and $\|\eta_C\|_2<\kappa_C\|g_C\|_2$, then
\begin{equation}
\left\|
\operatorname{unit}(v_C^{(\ell_c)})
-
\operatorname{unit}(g_C)
\right\|_2
\leq
\frac{2\|\eta_C\|_2}{\kappa_C\|g_C\|_2}.
\label{eq:method-recovery-bound}
\end{equation}
Thus, derivation is directionally stable when the pooled residual is small relative to the shared concept component. Appendix~\ref{app:theory-recovery} gives the derivation and evaluates this premise using held-out single-token concepts.

The carrier representation induces the lens effect
$e_C^{(\ell_c)}=J_{\ell_c}v_C^{(\ell_c)}$.
For a source layer $\ell_s\neq\ell_c$, we reconstruct a source-layer representation whose J-lens effect matches this carrier effect:
\begin{equation}
v_C^{(\ell_s)}
=
\operatorname*{arg\,min}_{v}
\left\{
\left\|
J_{\ell_s}v-e_C^{(\ell_c)}
\right\|_2^2
+
\lambda_{\ell_s}^{\mathrm{ridge}}\|v\|_2^2
\right\}.
\label{eq:transport}
\end{equation}
This transport has a direct spectral interpretation. If
$J_{\ell_s}=U\operatorname{diag}(\sigma_i)V^\top$, then the transported effect satisfies
\begin{equation}
\widehat e_C^{(\ell_s)}
=
J_{\ell_s}v_C^{(\ell_s)}
=
P_{\ell_s}e_C^{(\ell_c)},
\qquad
P_{\ell_s}
=
U\operatorname{diag}\!\left(
\frac{\sigma_i^2}
{\sigma_i^2+\lambda_{\ell_s}^{\mathrm{ridge}}}
\right)U^\top.
\label{eq:method-spectral-filter}
\end{equation}
Hence ridge transport preserves effect directions that are well represented by the source lens while attenuating poorly conditioned directions. Appendix~\ref{app:theory-transport} further separates the resulting discrepancy into carrier-derivation error and ridge bias.

Let $q_{\ell_s}=N(J_{\ell_s}h_{\ell_s})$ be the source query. We score each candidate by directional agreement between its transported effect and $q_{\ell_s}$:
\begin{equation}
G_C(h_{\ell_s})
=
m_U
\left\langle
\operatorname{unit}\!\left(
N(J_{\ell_s}v_C^{(\ell_s)})
\right),
q_{\ell_s}
\right\rangle.
\label{eq:concept-score}
\end{equation}
Here $N$ is the final normalization from Section~\ref{sec:method-setup}, $\operatorname{unit}(z)=z/\|z\|_2$, and $m_U$ is the median norm of the unembedding vectors. Pooling and ridge reconstruction do not preserve a meaningful absolute scale, so \textbf{SCORE} compares directions after normalization. The common factor $m_U$ places concept and vocabulary scores on the same numerical scale. Each vocabulary token $w$ is scored by the corresponding normalized rule
\[
m_U
\left\langle
\operatorname{unit}(u_w),
q_{\ell_s}
\right\rangle,
\]
so concepts and vocabulary tokens compete against the same query. This is also the scoring rule used for Template Lens directions in our comparison (Appendix~\ref{app:template}). The native J-lens ranking in Equation~\ref{eq:j-lens-readout}, which retains unembedding-row norms, is used to select first-token seeds for \textbf{PROPOSE}.

The normalization also clarifies which transport errors can affect the final score. Let
$x=N(e_C^{(\ell_c)})$, $\widehat x=N(\widehat e_C^{(\ell_s)})$,
$p=x/\|x\|_2$, and decompose the transport perturbation as
$\widehat x-x=ap+b$ with $b\perp p$. When
$\|\widehat x-x\|_2\leq\|x\|_2/2$,
\begin{equation}
\frac{G_C(h_{\ell_s})}{m_U}
=
q_{\ell_s}^{\top}p
+
\frac{q_{\ell_s}^{\top}b}{\|x\|_2}
+
\varepsilon,
\qquad
|\varepsilon|
\leq
4\|q_{\ell_s}\|_2
\frac{\|\widehat x-x\|_2^2}{\|x\|_2^2}.
\label{eq:method-score-stability}
\end{equation}
The radial component $ap$ therefore has no first-order effect after normalization; the leading score change is determined by the component $b$ that changes direction and is visible to the source query. Appendix~\ref{app:theory-score} gives the full finite-error analysis and the corresponding ranking stability result.

\section{Analysis of Concept Lens}
\label{sec:diagnosis}

We examine the first-token clues used for candidate generation and the concept information preserved in carrier representations. We evaluate instruction-tuned variants of Llama-3.1-8B, Gemma-3-12B, Qwen3-14B, Qwen3.6-27B, and Llama-3.3-70B~\citep{llama3,gemma3,qwen3}. For readability, we omit instruction-tuning suffixes from model names in the text.

\subsection{Can First-Token Clues Guide Concept Completion?}
\label{sec:first-token}

We measure first-token readout across each model's \emph{workspace band}, a middle-layer interval identified using geometric and persistence diagnostics and fixed before cloze evaluation (Appendix~\ref{app:workspace}). \textit{First-token Top-10} is the fraction of multi-token examples whose first token appears in the J-lens top-10 at any layer in the band. \textit{Single-token Top-10} measures the corresponding rate for single-token concepts. We stratify both measures by concept length.

For concepts containing two to four tokens under each model's tokenizer, we supply the correct first token and source cloze context, then let the model generate the continuation freely. \textit{Continuation Derivation} measures the fraction whose continuation exactly matches all remaining target tokens.

\begin{figure}[!ht]
\centering
\includegraphics[width=0.86\linewidth]{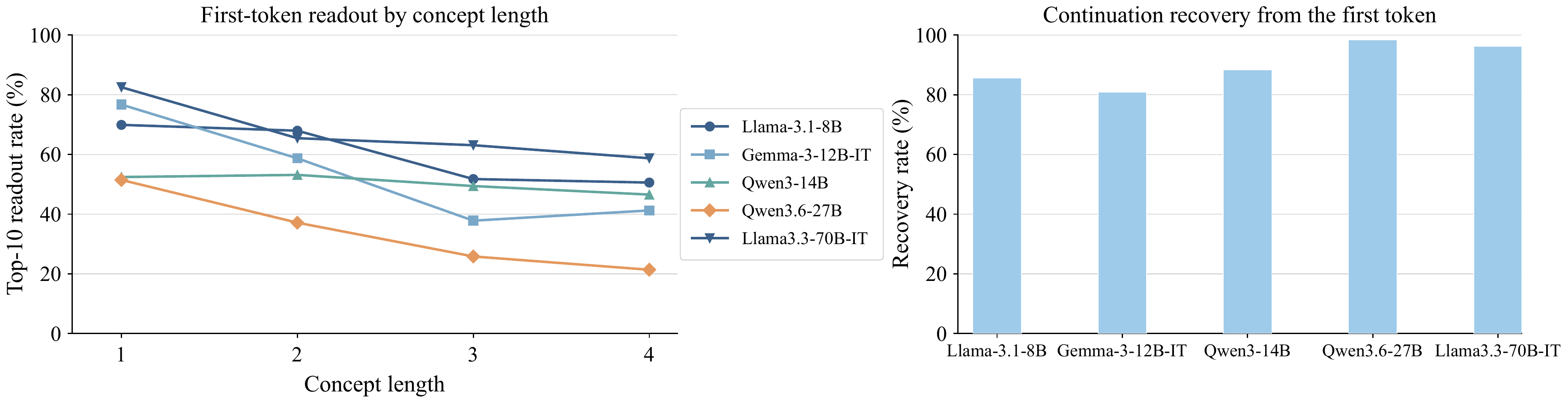}
\caption{\textbf{First-token readout and continuation derivation across five models.} \textit{Left:} Top-10 readout rate by concept length, aggregated across each model's workspace band. For single-token concepts, we use 103 single-token clozes from the official J-lens multi-hop cloze benchmark and evaluate the target token itself; for two- to four-token concepts, we evaluate whether the first target token appears in the J-lens top-10 within the band. Concept length is measured with each model's tokenizer, while the x-axis uses the common labels 1--4. \textit{Right:} Continuation derivation from the first token for two- to four-token concepts. The correct first token is provided with the source cloze context, and derivation succeeds only when all remaining target tokens are derived.}
\label{fig:first-token}
\end{figure}

Figure~\ref{fig:first-token} shows that first-token clues remain readable across concept lengths and that the frozen model completes concepts given the correct clue and source context.

\subsection{Do derived Representations Preserve Concept Identity?}
\label{sec:representation-recovery}

Our held-out set contains 500 words that are single tokens in all five models (Appendix~\ref{app:data-calibration}). Each word's native J-lens direction serves as the reference. We split each word into two tokens (e.g., \textit{airport} $\rightarrow$ \textit{air}+\textit{port}), derive $v_C$ from the fragments, and rank the 500 reference directions by cosine similarity. We report \textit{Top-1} accuracy and \textit{Wrong Similarity}, the mean cosine similarity to incorrect references. Baselines use the first fragment, the mean of both fragments, or a random representation.

The derived representations achieve 98.0--99.8\% Top-1 accuracy, compared with at most 66.8\% for fragment-based baselines, and substantially lower \textit{Wrong Similarity} (Table~\ref{tab:vector-recovery}).

\begin{table}[!ht]
\caption{\textbf{derived representations preserve concept identity.} We split held-out single-token words into two-token forms and retrieve each word's native J-lens direction from a 500-word reference bank. Top-1 (\%) measures retrieval accuracy; Wrong Sim. is the mean cosine similarity to incorrect reference directions. Higher Top-1 and lower Wrong Sim. indicate better preservation of concept identity.}
\label{tab:vector-recovery}
\centering
\scriptsize
\renewcommand{\arraystretch}{1.08}
\setlength{\tabcolsep}{2.2pt}
\resizebox{\linewidth}{!}{%
\begin{tabular}{@{}l*{10}{r}@{}}
\toprule
& \multicolumn{2}{c}{\textbf{Llama-3.1-8B}}
& \multicolumn{2}{c}{\textbf{Gemma-3-12B}}
& \multicolumn{2}{c}{\textbf{Qwen3-14B}}
& \multicolumn{2}{c}{\textbf{Qwen3.6-27B}}
& \multicolumn{2}{c}{\textbf{Llama-3.3-70B}} \\
\cmidrule(lr){2-3} \cmidrule(lr){4-5} \cmidrule(lr){6-7} \cmidrule(lr){8-9} \cmidrule(lr){10-11}
\textbf{Method}
& \multicolumn{1}{c}{\textbf{Top-1} $\uparrow$}
& \multicolumn{1}{c}{\textbf{Wrong Sim.} $\downarrow$}
& \multicolumn{1}{c}{\textbf{Top-1} $\uparrow$}
& \multicolumn{1}{c}{\textbf{Wrong Sim.} $\downarrow$}
& \multicolumn{1}{c}{\textbf{Top-1} $\uparrow$}
& \multicolumn{1}{c}{\textbf{Wrong Sim.} $\downarrow$}
& \multicolumn{1}{c}{\textbf{Top-1} $\uparrow$}
& \multicolumn{1}{c}{\textbf{Wrong Sim.} $\downarrow$}
& \multicolumn{1}{c}{\textbf{Top-1} $\uparrow$}
& \multicolumn{1}{c}{\textbf{Wrong Sim.} $\downarrow$} \\
\midrule
\(\mathrm{Ours}\)
& 99.0 & -0.0096
& 98.0 & 0.0024
& 99.2 & -0.0015
& 99.4 & -0.0088
& 99.8 & 0.0114 \\
\addlinespace[1pt]
\(\mathrm{First}\)
& 55.6 & 0.0675
& 29.0 & 0.1177
& 50.0 & 0.0501
& 58.2 & 0.0649
& 48.2 & 0.0827 \\
\(\mathrm{Mean}\)
& 66.0 & 0.0699
& 23.4 & 0.1246
& 52.8 & 0.0564
& 66.8 & 0.0757
& 52.4 & 0.0886 \\
\addlinespace[1pt]
\(\mathrm{Random}\)
& 0.2 & 0.0001
& 0.2 & 0.0004
& 0.0 & -0.0000
& 0.0 & 0.0001
& 0.4 & 0.0001 \\
\bottomrule
\end{tabular}
}
\end{table}
\FloatBarrier

\begin{keybox}{Key Finding}
First-token clues remain readable across concept lengths, and the model completes concepts when given the correct clue and source context. Carrier states yield representations that preserve concept identity in the controlled reconstruction task.
\end{keybox}

\section{Experiments}
\label{sec:experiments}

Section~\ref{sec:diagnosis} examines the two stages of Concept Lens. We now evaluate the complete method with J-lens and R-lens across the same five models. We test whether Concept Lens identifies latent intermediate multi-token concepts (Section~\ref{sec:readout}) and whether the derived representations can redirect model computation (Section~\ref{sec:exp-swap}). Appendix~\ref{app:lens-extension} defines the common interface used for the two lenses.

\subsection{End-to-End Readout}
\label{sec:readout}

We evaluate the pipeline on $2{,}400$ directed clozes from $1{,}200$ counterfactual pairs across five models. Appendix~\ref{app:data} describes the benchmark construction, composition, and data splits. Each cloze refers to its intermediate multi-token concept only through indirect clues. We report \textbf{Coverage}, the fraction of clozes whose target is proposed at any layer within the model-specific workspace band, and \textbf{Rank@10}, the fraction whose target is ranked in the top-10 at any layer within that band. Both metrics are means of per-cloze success indicators. Following the J-lens evaluation protocol, we use the pre-specified workspace band, fixed before task evaluation, and run our \textbf{PROPOSE}/\textbf{SCORE} pipeline across the layers in that band (Appendix~\ref{app:workspace}).

We use two settings to distinguish candidate availability from scoring. In \emph{end-to-end}, the pipeline must propose and then rank the target. In the \emph{oracle-candidate setting}, we add the annotated target to each layer's candidate set to evaluate \textbf{SCORE} independently of whether \textbf{PROPOSE} finds the target. Comparing the settings measures how target availability affects end-to-end readout. The \emph{Random} reference reuses the J-lens candidate bank, uniformly and independently permutes the joint ranking of vocabulary tokens and available candidates at each layer, and applies the same workspace-band aggregation, so its Coverage matches J-lens and its Rank@10 values are analytic expectations.

Table~\ref{tab:ranking} shows that \textbf{SCORE} ranks the target in the top-10 for 80.2--98.8\% of clozes when the target is added to the candidate set, compared with 23.1--61.9\% for the complete pipeline. End-to-end Rank@10 closely tracks Coverage while oracle Rank@10 remains high, identifying candidate coverage as a major bottleneck. The ranking also depends on the source hidden state. With the candidate set and derived directions held fixed, changing only the source state sharply changes concept selection across counterfactual inputs (Appendix~\ref{app:activation_readout}).

The procedure successfully generalizes across different readout types. Both J-lens and R-lens outperform Template Lens, whose predefined concept list covers 25.7\% of targets and whose end-to-end Rank@10 is only 19.9--23.5\%. We give Template Lens a fixed bank of 38,400 concepts, three times the 12,800-concept configuration, to make this coverage comparison conservative (Appendix~\ref{app:template}). R-lens gains 15.2--24.2 points over J-lens in end-to-end Rank@10, consistent with the stronger early-layer readouts reported for R-lens~\citep{blank2026rlens}. The end-to-end J/R difference combines changes in their seed rankings and scoring geometries; the oracle-candidate setting evaluates the ranking stage once the target is available.

\begin{table*}[t]
\caption{\textbf{Multi-token concept readout across five models.} Across $2{,}400$ clozes per model, \textbf{Coverage} measures target availability and \textbf{Rank@10} measures successful top-10 retrieval at any layer within the model-specific workspace band. \emph{End-to-end} evaluates the full pipeline; the \emph{Oracle-candidate setting} adds the target to each layer's candidate set to isolate \textbf{SCORE} from \textbf{PROPOSE}.}
\label{tab:ranking}
\centering
\small
\renewcommand{\arraystretch}{1.08}
\setlength{\tabcolsep}{2.4pt}
\resizebox{\textwidth}{!}{%
\begin{tabular}{@{}l*{10}{r}@{}}
\toprule
& \multicolumn{2}{c}{\textbf{Llama-3.1-8B}} & \multicolumn{2}{c}{\textbf{Gemma-3-12B}} & \multicolumn{2}{c}{\textbf{Qwen3-14B}} & \multicolumn{2}{c}{\textbf{Qwen3.6-27B}} & \multicolumn{2}{c}{\textbf{Llama-3.3-70B}} \\
\cmidrule(lr){2-3} \cmidrule(lr){4-5} \cmidrule(lr){6-7} \cmidrule(lr){8-9} \cmidrule(lr){10-11}
\textbf{Method} & \multicolumn{1}{c}{\textbf{Coverage}} & \multicolumn{1}{c}{\textbf{Rank@10} $\uparrow$} & \multicolumn{1}{c}{\textbf{Coverage}} & \multicolumn{1}{c}{\textbf{Rank@10} $\uparrow$} & \multicolumn{1}{c}{\textbf{Coverage}} & \multicolumn{1}{c}{\textbf{Rank@10} $\uparrow$} & \multicolumn{1}{c}{\textbf{Coverage}} & \multicolumn{1}{c}{\textbf{Rank@10} $\uparrow$} & \multicolumn{1}{c}{\textbf{Coverage}} & \multicolumn{1}{c}{\textbf{Rank@10} $\uparrow$} \\
\midrule  
\multicolumn{11}{@{}l}{\textit{End-to-end: Proposal + Scoring}} \\
\addlinespace[1pt]
\rowcolor{gray!10} \(\mathrm{Ours\ (J\text{-}lens)}\) & \underline{49.2} & \underline{44.3} & \underline{40.4} & \underline{37.4} & \underline{41.6} & \underline{31.5} & \underline{30.1} & \underline{23.1} & \underline{56.4} & \underline{46.5} \\
\rowcolor{gray!10} \(\mathrm{Ours\ (R\text{-}lens)}\) & \textbf{63.6} & \textbf{61.9} & \textbf{60.8} & \textbf{52.6} & \textbf{65.6} & \textbf{55.7} & \textbf{43.0} & \textbf{40.3} & \textbf{75.5} & \textbf{61.8} \\
\(\mathrm{Template\ Lens}\) & 25.7 & 19.9 & 25.7 & 22.2 & 25.7 & 21.0 & 25.7 & 21.9 & 25.7 & 23.5 \\
\(\mathrm{Random}\) & 49.2 & 0.03 & 40.4 & 0.01 & 41.6 & 0.01 & 30.1 & 0.0 & 56.4 & 0.08 \\
\midrule
\addlinespace[3pt]
\multicolumn{11}{@{}l}{\textit{Oracle-candidate setting: Scoring only}} \\
\addlinespace[1pt]
\rowcolor{gray!10} \(\mathrm{Ours\ (J\text{-}lens)}\) & 100.0 & \underline{91.9} & 100.0 & \textbf{90.9} & 100.0 & \textbf{98.8} & 100.0 & \underline{80.2} & 100.0 & \textbf{83.3} \\
\rowcolor{gray!10} \(\mathrm{Ours\ (R\text{-}lens)}\) & 100.0 & \textbf{95.9} & 100.0 & \underline{90.6} & 100.0 & \underline{90.5} & 100.0 & \textbf{90.6} & 100.0 & \underline{81.8} \\
\(\mathrm{Template\ Lens}\) & 100.0 & 67.5 & 100.0 & 69.9 & 100.0 & 73.2 & 100.0 & 78.6 & 100.0 & 80.8 \\
\(\mathrm{Random}\) & 100.0 & 0.20 & 100.0 & 0.14 & 100.0 & 0.18 & 100.0 & 0.24 & 100.0 & 0.33 \\
\bottomrule
\end{tabular}%
}
\end{table*}


\subsection{Causal Concept Swap}
\label{sec:exp-swap}

We test whether derived representations can redirect model computation on paired clozes with the same relation. The source cloze requires $C$, and its partner requires $D$. \textbf{SCORE} derives representations of these annotated concepts, which we transport to fixed intervention layers. \emph{Full replacement} removes the $C$ component from the source activation and adds the $D$ direction. We keep the source prompt unchanged and test whether the answer shifts toward the partner answer. Addition strength is measured on the clean partner activation, without fitting to intervention outcomes. \emph{Add-only} and \emph{delete-only} controls isolate the two operations; magnitude-matched random directions test specificity. We evaluate both directions of each pair, yielding 2,400 directed trials. Appendix~\ref{app:causal} gives the full protocol.

\begin{table*}[htbp]
\caption{\textbf{derived concept representations causally redirect intermediate computation.}
Controlled swaps on paired clozes requiring different concepts. All metrics are percentages of trials.
\textbf{P@10}: partner answer appears in the edited top-10; \textbf{S@10}: source answer remains in the edited top-10; \textbf{CS@10}: partner ranks in the top-10 and source ranks outside the top-10.
\emph{Full replacement} removes the source direction and adds the transported partner direction; \emph{add-only} and \emph{delete-only} isolate the two operations.
\emph{Random} uses magnitude-matched directions in the add-only setting.
}
\label{tab:swap}
\centering
\small
\renewcommand{\arraystretch}{1.10}
\setlength{\tabcolsep}{1.7pt}
\resizebox{\textwidth}{!}{%
\begin{tabular}{@{}l*{15}{r}@{}}
\toprule
& \multicolumn{3}{c}{\textbf{Llama-3.1-8B}}
& \multicolumn{3}{c}{\textbf{Gemma-3-12B}}
& \multicolumn{3}{c}{\textbf{Qwen3-14B}}
& \multicolumn{3}{c}{\textbf{Qwen3.6-27B}}
& \multicolumn{3}{c}{\textbf{Llama-3.3-70B}} \\
\cmidrule(lr){2-4} \cmidrule(lr){5-7} \cmidrule(lr){8-10} \cmidrule(lr){11-13} \cmidrule(lr){14-16}
\textbf{Method}
& \multicolumn{1}{c}{\textbf{P@10} $\uparrow$}
& \multicolumn{1}{c}{\textbf{S@10} $\downarrow$}
& \multicolumn{1}{c}{\textbf{CS@10} $\uparrow$}
& \multicolumn{1}{c}{\textbf{P@10} $\uparrow$}
& \multicolumn{1}{c}{\textbf{S@10} $\downarrow$}
& \multicolumn{1}{c}{\textbf{CS@10} $\uparrow$}
& \multicolumn{1}{c}{\textbf{P@10} $\uparrow$}
& \multicolumn{1}{c}{\textbf{S@10} $\downarrow$}
& \multicolumn{1}{c}{\textbf{CS@10} $\uparrow$}
& \multicolumn{1}{c}{\textbf{P@10} $\uparrow$}
& \multicolumn{1}{c}{\textbf{S@10} $\downarrow$}
& \multicolumn{1}{c}{\textbf{CS@10} $\uparrow$}
& \multicolumn{1}{c}{\textbf{P@10} $\uparrow$}
& \multicolumn{1}{c}{\textbf{S@10} $\downarrow$}
& \multicolumn{1}{c}{\textbf{CS@10} $\uparrow$} \\
\midrule
\multicolumn{16}{@{}l}{\textit{Full replacement}} \\
\rowcolor{gray!10}\(\mathrm{Ours\ (J\text{-}lens)}\) & 45.5 & 17.2 & 38.2 & 48.0 & 11.8 & 44.2 & 48.6 & 16.2 & 41.9 & 27.6 & 31.8 & 15.7 & 60.2 & 3.8 & 59.0 \\
\rowcolor{gray!10}\(\mathrm{Ours\ (R\text{-}lens)}\) & 50.0 & 14.0 & 43.8 & 46.4 & 13.4 & 42.0 & 50.0 & 16.5 & 43.2 & 30.6 & 29.5 & 18.9 & 60.9 & 4.3 & 59.4 \\
\(\mathrm{Template\ Lens}\) & 27.9 & 81.1 & 7.2 & 12.2 & 60.8 & 4.3 & 14.8 & 91.6 & 2.1 & 6.7 & 50.0 & 1.4 & 16.8 & 41.0 & 9.8 \\
\(\mathrm{Random}\) & 3.6 & 89.6 & 1.6 & 3.1 & 93.1 & 1.4 & 4.0 & 97.4 & 1.5 & 4.8 & 96.8 & 1.1 & 1.7 & 92.7 & 0.8 \\
\midrule
\addlinespace[3pt]
\multicolumn{16}{@{}l}{\textit{Operation controls}}\\
\addlinespace[1pt]
\(\mathrm{Ours\ (J\text{-}lens),\ add\ only}\) & 39.6 & 38.7 & 22.2 & 40.9 & 26.2 & 31.8 & 42.8 & 38.6 & 24.8 & 20.1 & 48.7 & 6.8 & 58.0 & 8.3 & 54.9 \\
\(\mathrm{Ours\ (J\text{-}lens),\ delete\ only}\) & 4.0 & 33.1 & 1.7 & 3.9 & 31.8 & 1.1 & 4.5 & 30.0 & 1.0 & 4.8 & 43.2 & 1.2 & 5.4 & 41.7 & 1.5 \\
\(\mathrm{Ours\ (R\text{-}lens),\ add\ only}\) & 44.3 & 35.1 & 28.2 & 38.8 & 28.2 & 29.0 & 45.0 & 38.1 & 26.8 & 22.8 & 47.6 & 9.0 & 58.3 & 8.8 & 54.9 \\
\(\mathrm{Ours\ (R\text{-}lens),\ delete\ only}\) & 3.7 & 32.5 & 1.4 & 3.7 & 33.1 & 0.9 & 4.4 & 29.6 & 1.0 & 4.9 & 42.8 & 1.2 & 5.3 & 42.6 & 1.5 \\
\bottomrule
\end{tabular}
}
\end{table*}

Concept Lens promotes the partner answer in 27.6--60.9\% of trials and achieves 15.7--59.4\% CS@10, compared with 1.4--9.8\% CS@10 for Template Lens (Table~\ref{tab:swap}). Full replacement yields the highest clean-swap rate among the operation controls across all five models and both underlying lenses. Add-only promotes the partner answer but leaves more source answers intact, whereas delete-only suppresses the source answer but rarely promotes the partner answer (3.7--5.4\% P@10). Random directions yield few clean swaps, supporting the specificity of the derived directions.

\section{Applications}
\label{sec:applications}

The controlled benchmark provides annotated intermediate concepts for quantitative evaluation. We next use Concept Lens to compare model interpretations across matched inputs in three settings spanning logical consistency, problem structure, and pragmatic interpretation.

\begin{figure}[htbp]
\centering
\includegraphics[width=0.96\linewidth]{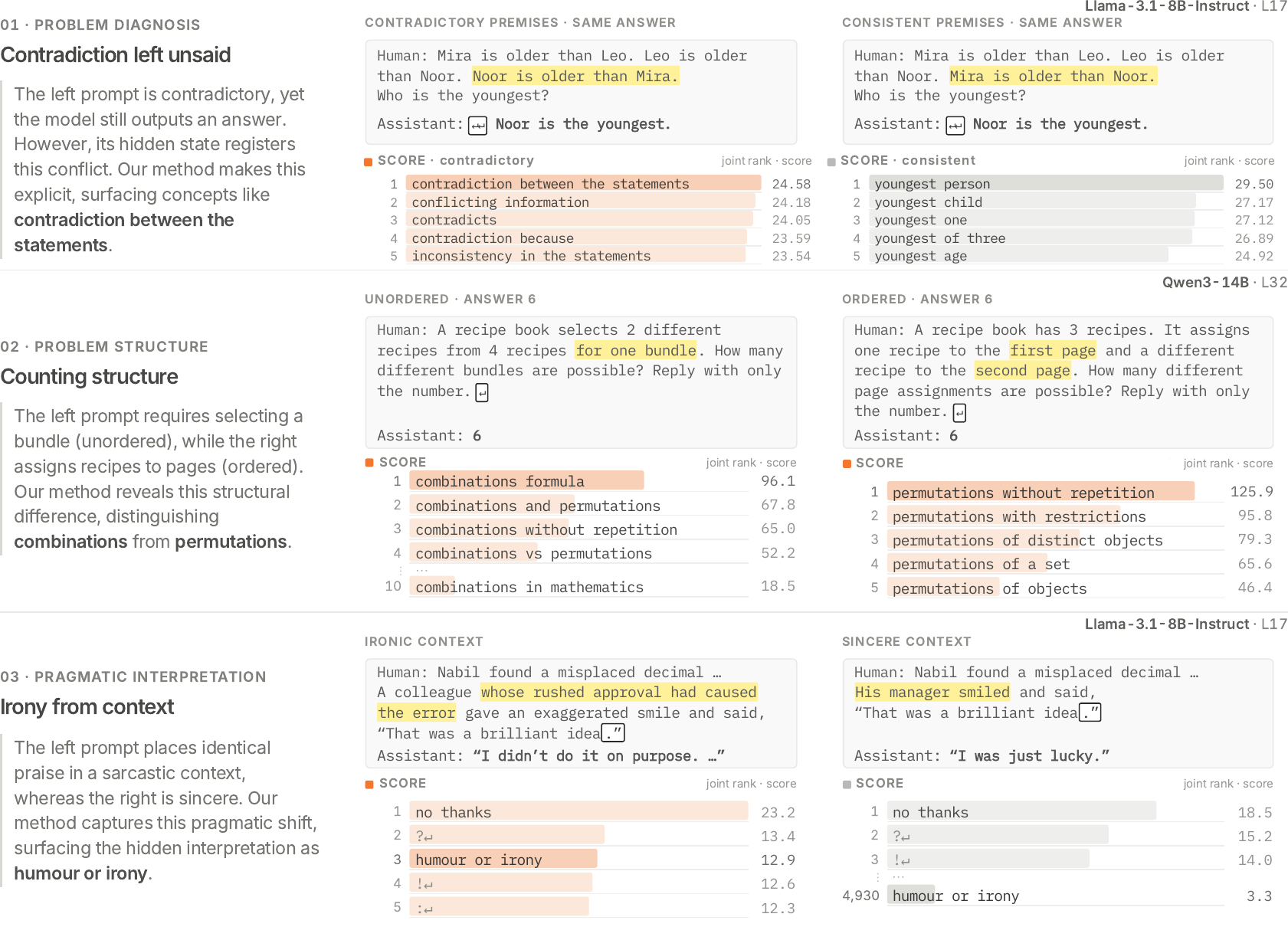}
\caption{\textbf{Concept Lens reveals internal interpretations behind similar final responses.} The top two pairs produce identical answers, while the bottom pair gives similar replies with different pragmatic meanings. The panels show latent contradictions (top), counting structures (middle), and pragmatic interpretation (bottom). The top and bottom examples use Llama-3.1-8B at L17; the middle uses Qwen3-14B at L32. Bar charts display the top-ranked concepts extracted at the boxed token; yellow highlights mark the context variations.}
\label{fig:applications}
\end{figure}

Figure~\ref{fig:applications} shows these distinctions. In the first pair, both versions yield the same answer ``Noor is the youngest,'' yet the
source states differ: the contradictory context shifts the readout toward \emph{contradiction between the statements}, while the consistent context favors \emph{youngest person}. In the second, both prompts yield the answer 6, yet the source states differ: one is associated with \emph{combinations}, the other with \emph{permutations}, exposing different reasoning structures behind the shared final output. In the third, the same surface phrase is interpreted differently under different contexts: \emph{humour or irony} ranks 3rd in the sarcastic context but 4,930th in the sincere one, showing that the source state encodes pragmatic interpretation beyond what the phrase itself determines. Thus, multi-token readout distinguishes internal interpretations that remain implicit in similar final responses.

\begin{keybox}{Implication}
Multi-token readout provides a more complete conceptual view than isolated token-level readouts, making otherwise hidden distinctions easier to inspect.
\end{keybox}

\newpage
\section{Related Work}
\label{sec:related}

Concept Lens connects lens-based concept scoring with language-model interpretations of activations. Template and Oracle Lens associate concepts with representations for scoring and intervention (Appendix~\ref{app:related-template}). Patchscopes and SelfIE generate multi-token descriptions from injected activations~\citep{pmlr-v235-ghandeharioun24a,pmlr-v235-chen24ao}. Recent work identifies linguistic-prior bias in Patchscopes~\citep{gong2026faithfulpatchscopes} and sensitivity to prompts and injection settings in related introspection tasks~\citep{lindsey2026emergent,singh2026realitycheck}. Natural language autoencoders connect generated descriptions to activation reconstruction~\citep{frasertaliente2026nla}. Concept Lens separates candidate generation from source-conditioned scoring, with reliability assessed through representation derivation and fixed-candidate source-state swaps. Studies of lexical representations and latent reasoning motivate complete concepts as readout units (Appendix~\ref{app:related-scope}). Appendix~\ref{app:related} expands these comparisons and discusses feature dictionaries and activation steering.

\section{Discussion and Limitation}
\label{sec:discussion}

Controlled comparisons show that Concept Lens selects concepts according to the hidden state being inspected. With candidates and representations fixed, replacing the source state shifts the ranking toward the paired concept (Appendix~\ref{app:activation_readout:mismatch}). On fixed candidate banks containing naturally proposed targets, \textbf{SCORE} achieves higher Rank@10 than continuation likelihood across five models (Appendix~\ref{app:activation_readout:scoring}). The readout also distinguishes concepts sharing a first token (Appendix~\ref{app:activation_readout:first-token}), supporting its ability to resolve complete concepts. The causal swaps complement these readout tests by evaluating the behavioral effects of the concept representations. Together, the experiments connect concept selection to the hidden state being inspected and to the model's subsequent computation.

The gap between end-to-end and oracle-candidate results identifies target coverage as a major bottleneck. Appendix~\ref{app:activation_readout:proposal} compares two proposal strategies using the same scoring procedure, showing how candidate generation affects retrieval. Higher target coverage at a fixed candidate budget offers a route to stronger readout with the same carrier reader and scoring rule. The proposal scaffold can also guide candidate generation toward concepts relevant to a particular analysis.
Concept Lens reuses the underlying LLM and calibrates a single carrier reader across multiple layers. Given an existing lens, calibration on Qwen3.6-27B, for example, takes only $0.1$ A100-hours for all 52 layers. Template Lens spends $20$--$30$ A100-hours plus \$1{,}500--\$2{,}000 in API costs to construct its concept bank. Oracle Lens preparation is estimated to require $2{,}000$--$4{,}000$ A100-hours for a single-layer configuration. Concept Lens constructs candidate representations at query time and reads all 52 layers in about $4$ seconds (Appendix~\ref{app:cost}).

\subsection{Limitations}
\label{sec:limitations}

Candidate coverage limits end-to-end concept readout. Stronger proposers could improve coverage while retaining the same carrier reader and scoring rule. Our direct comparison with Oracle Lens covers the public Qwen3.6-27B configuration at one shared activation layer (Appendix~\ref{app:oracle}). A broader comparison could evaluate both methods over matched layer ranges across several models. Tracking concept derivation across depth would show where each readout is most effective, while reporting preparation costs for each configuration would connect readout quality with adaptation cost. Ground truth remains a broader challenge for activation readouts, as the intermediate concepts implied by our cloze tasks are not guaranteed to be represented in the model’s computation. Our controlled split-token test provides a well-defined reference for representation derivation.

\section{Conclusion}
\label{sec:conclusion}

Concept Lens reads latent concepts from LLM activation with lightweight adaptation. The method uses the frozen model to propose complete concepts and scores their derived representations in an existing lens readout space. Experiments across five models demonstrate effective concept retrieval, while causal interventions support the connection between the readout and subsequent model behavior. The application cases reveal different internal interpretations behind similar final answers, illustrating how concept readout makes implicit aspects of model computation accessible.

\newpage

\subsection*{AI use statement}

In this work, we used generative AI tools for generating and processing parts of the synthetic dataset. The authors determined the intermediate concepts, relations, templates, and generation constraints, and generative AI was used to produce candidate clue sentences under these constraints. We also used generative AI tools for literature search, language editing, and assistance with drafting methodological explanations, comparison protocols, and supporting mathematical analysis. AI assistance included checking implementation details and aggregating saved experimental results. All AI-generated dataset content was subsequently reviewed and processed by the authors. All code and data used for the reported results were checked by at least two authors. We have reviewed all AI-assisted work and take responsibility for the final content of this work, including text, claims, code, data, and experimental results.

\subsection*{Reproducibility statement}

We have made reproducibility a central part of this work. We will release all code, datasets, configuration files, and implementation details required to reproduce our results through an anonymous repository, and we will provide the exact sources and versions of all model weights and external resources used in our experiments. The appendix describes the algorithms, data construction procedure, evaluation protocol, hyperparameters, and other implementation details needed to reproduce the experiments. We will also release the processed datasets and relevant intermediate artifacts so that individual stages of the pipeline can be checked independently. All experiments in this paper can be reproduced with approximately 500 A100 GPU-hours of computation. We have aimed to make every reported result independently reproducible and to minimize differences caused by missing implementation details or undocumented experimental settings.

\bibliography{references}
\bibliographystyle{iclr2027_conference}

\appendix

\input{appendix/appendix}

\end{document}